\documentclass[10pt]{article}
\usepackage{rotating, epsfig}
\usepackage[english]{babel}
\usepackage{amsfonts,amsmath}
\usepackage{color}
\usepackage[utf8x]{inputenc}
\usepackage[cyr]{aeguill}

\newcommand{\MC}{\mathcal{C}}

\newcommand{\BSN}{\boldsymbol N}

\newcommand{\BSX}{\boldsymbol X}

\newcommand{\bsn}{\boldsymbol n}

\newcommand{\bsx}{\boldsymbol x}

\newcommand{\ind}{\mbox{\rm 1\kern-.4em 1}}

\newcommand{\pa}{\mbox{pa}}
\newcommand{\spa}{\mbox{\footnotesize{pa}}}

\title{Parametric Modelling of Multivariate Count Data 
Using Probabilistic~Graphical~Models}

{\author{Pierre Fernique \thanks{CIRAD, UMR AGAP and Inria, Virtual
Plants, Montpellier, France} \and Jean-Baptiste Durand
\thanks{Univ. Grenoble Alpes, Laboratoire Jean Kuntzmann and Inria,
  Mistis, 51 rue des Math\'ematiques, B.P. 53, F-38041 Grenoble Cedex
  9, France.
Email~:~{\texttt{jean-baptiste.durand@imag.fr}}} \and Yann
Gu\'edon${}^{\, *}$}}

\date{}

\begin{document}


\maketitle 

\section{Introduction}
\label{sec:intro}

Multivariate count data are defined as the number of items of different
categories issued from sampling within a population, which individuals
are grouped into categories. The analysis of multivariate count data
is a recurrent and crucial issue in numerous modelling problems,
particularly in the fields of biology and ecology (where the data can
represent, for example, children counts associated with multitype
branching processes), sociology and econometrics. Denoting by $K$ the
number of categories, multivariate 
count data analysis relies on modelling the joint distribution 
of the $K$-dimensional random vector $\BSN=(N_0,\ldots,N_{K-1})$ with
discrete components. We focus on I) Identifying categories that appear
simultaneously, or on the contrary that are mutually
exclusive. This is achieved 
by identifying conditional independence relationships
between the $K$ variables; II)Building parsimonious parametric
models consistent with these relationships; III) Characterising and
testing the effects of covariates on the distribution of $\BSN$,
particularly on the dependencies between its components.

To achieve these goals, we propose an approach based on graphical
probabilistic models (Koller \& Friedman, 2009\nocite{Koller09}) to
represent the conditional independence relationships in $\BSN$, 
and on parametric distributions to ensure model parsimony. Three kinds of
graphs are usually considered: either undirected (UG), directed
acyclic (DAG), or partially directed acyclic (PDAG). Models and
methods for graph 
identification were proposed in UGs, using frequencies
to estimate the probabilities (so-called {\it nonparametric
 estimation}), using mutual information -- see Meyer {\it et al.}
(2008\nocite{Meyer08}) and references therein. Under a multivariate Gaussian
assumption, an approach based on a L1 penalisation (lasso) was proposed
by Friedman {\it et al.} (2008\nocite{Friedman08}), with some
extension to Poisson distributions with log-linear models 
for dependencies (Allen \& Liu, 2012\nocite{Allen12}). Specific models
and methods were developed for DAGs. Most methods for graph
identification in DAGs are based on exploring the set of possible graphs using
some heuristic (e.g. hill climbing) and by scoring the visited graphs
(e.g. using BIC), the graph with highest score being eventually
selected -- see Koller \& Friedman (2009\nocite{Koller09}) 
for a review. The case of parametric models for PDAGs has been
considered less often in the literature. A family of such models was
proposed by Johnson \& Hoeting (2011\nocite{Johnson11}) using conditional
Gaussian distributions, but the problem of graph identification was
not addressed. Lee \& Hastie (2012\nocite{Lee12}) addressed the
problem of graph identification in graphical models with both
continuous and discrete random variables, but in a restrictive setting
of UGs with conditional Gaussian and multinomial distributions.

Our context of application is characterised by zero-inflated, often
right skewed marginal distributions. Thus, Gaussian and Poisson 
distributions are not {\it a priori} appropriate. Moreover, the
multivariate histograms typically have many cells, most of which 
are empty. Consequently, nonparametric estimation is not efficient.
 
\section{PDAG modelling and graph search}
\label{sec:model}
A family of parametric PDAG models is considered, such that covariates
can be introduced easily and in a flexible manner. The advantage of
PDAGs is that both marginal independence relationships and cyclic
dependencies between quadruplets of variables (at least) can be represented. 
The class of considered PDAGs is such that the joint distribution
factorises as 
\begin{equation}
P(\BSN=\bsn) = \prod\limits_{c \in \MC} 
P(\BSN_c=\bsn_c | \BSN_{\spa(c)}=\bsn_{\spa(c)}),
\label{eq:fact}
\end{equation}
where $\MC$ denotes the set of undirected subgraphs (so-called {\it
  chain components}), $\pa(c)$ the parent chain components of $c$
(which can be the empty set), and $\BSN_c$ denotes $\{N_j\}_{j \in
  c}$. We refer to Koller \& Friedman (2009\nocite{Koller09})
regarding graph terminology.

Each source vertex of the graph is associated with some univariate
distribution chosen among the binomial, negative binomial and Poisson
families and mixtures of such distributions. Each non-singleton source
component of the graph is associated with some multivariate
distribution chosen among diverse extensions of the multinomial
family, the multivariate Poisson distribution (Karlis
2003\nocite{Karlis03}) and mixtures of such distributions. Each
component of the graph with at least one parent is 
associated with the corresponding families of univariate and
multivariate regression models defined hereinbefore in the case of
source components. As a consequence, each 
factor in \eqref{eq:fact} is modelled by a parametric distribution or a
regression. The parameters are estimated by maximum likelihood, 
and the family with maximal BIC value is selected for each factor, so
that the joint distribution is uniquely defined. 

Graph search is achieved by a stepwise approach, issued from
unification of previous algorithms presented in Koller \& Friedman
(2009\nocite{Koller09}) for DAGs: Hill climbing, greedy
search, first ascent and simulated annealing. The search algorithm has
been improved by taking into account the parametric distribution
assumptions, which led to caching overlapping graphs at each step. 
An adaptation to PDAGs of graph search algorithms for DAGs has been
developed, by defining new operators: edge addition and deletion, and
directed edge reversal at chain component scale (instead of vertex
scale). Two operators specific to PDAGs have been added: chain
component addition and deletion. On the one hand, a parent vertex can
be added to its child chain component, or a child vertex can be added
to one of its parent chain component, which results into deletion of
one chain component in both cases. On the other hand, a vertex from a
chain component $c$ can be set to be a parent or a child of $c$, which
results into addition of one chain component. 

Since our model is essentially defined by chaining regression models
in PDAGs, some set of covariates $\BSX$ can be easily incorporated in
the model. This is achieved by substituting 
$P(\BSN=\bsn | \BSX=\bsx)$ for $P(\BSN=\bsn)$ in \eqref{eq:fact}, and 
$P(\BSN_c=\bsn_c | \BSN_{\spa(c)}=\bsn_{\spa(c)}, \BSX=\bsx)$ for 
$P(\BSN_c=\bsn_c | \BSN_{\spa(c)}=\bsn_{\spa(c)})$. In the graph
search step, some covariates in the set $\BSX$ may be discarded in
practice in $P(\BSN_c=\bsn_c | \BSN_{\spa(c)}=\bsn_{\spa(c)},
\BSX=\bsx)$, in a differentiated way with respect to the different chain
components $c$.

Comparisons between the different algorithms introduced in
Sections \ref{sec:intro} and \ref{sec:model} have been
performed on simulated datasets to:
(i) Assess gain in speed induced by caching;
(ii) Compare the graphs obtained under parametric and
  nonparametric distributions assumptions;
(iii) Compare different strategies for graph initialisation. 
  Strategies based on several random graphs have been compared to
  those based on a fast estimation of an UG, assumed to be the moral
  graph. This can be obtained by either applying the approach of Meyer {\it et
    al.} (2008\nocite{Meyer08}) or that of Friedman {\it et al.}
  (2008\nocite{Friedman08}), and then directing edges until a valid
  PDAG is obtained, using a modified version of Verma \& Pearl's
  (1992\nocite{Verma92}) DAG construction algorithm.

Similar comparisons (including between DAG and PDAG models) have also
been performed on real-world benchmark datasets issued from Chickering
(2002\nocite{chickering02a}). 
We also propose an original application of multitype branching
processes to the characterisation of growth patterns in Mango trees,
in relation to asynchrony.


\begin{thebibliography}{1}

\bibitem{Allen12}
{\sc Allen, G.~I., and Liu, Z.}
\newblock A log-linear graphical model for inferring genetic networks from
  high-throughput sequencing data.
\newblock In {\em Bioinformatics and Biomedicine (BIBM), 2012 IEEE
  International Conference on\/} (2012), IEEE, pp.~1--6.

\bibitem{chickering02a}
{\sc Chickering, D.}
\newblock Learning equivalence classes of bayesian-network structures.
\newblock {\em The Journal of Machine Learning Research 2\/} (2002), 445--498.

\bibitem{Friedman08}
{\sc Friedman, J., Hastie, T., and Tibshirani, R.}
\newblock Sparse inverse covariance estimation with the graphical lasso.
\newblock {\em Biostatistics 9}, 3 (2008), 432--441.

\bibitem{Johnson11}
{\sc Johnson, D., and Hoeting, J.}
\newblock Properties of graphical regression models for multidimensional
  categorical data.
\newblock {\em Statistics and Probability Letters 81}, 10 (2011), 1471--1475.

\bibitem{Karlis03}
{\sc Karlis, D.}
\newblock An {EM} algorithm for multivariate {P}oisson distribution and related
  models.
\newblock {\em Journal of Applied Statistics 30}, 1 (2003), 63--77.

\bibitem{Koller09}
{\sc Koller, D., and Friedman, N.}
\newblock {\em Probabilistic graphical models: principles and techniques}.
\newblock MIT press, 2009.

\bibitem{Lee12}
{\sc Lee, J.~D., and Hastie, T.~J.}
\newblock Structure learning of mixed graphical models.
\newblock Submitted to the Journal of Machine Learning Research. Available:
  http://www.stanford.edu/~hastie/Papers/structmgm.pdf, 2012.

\bibitem{Meyer08}
{\sc Meyer, P., Lafitte, F., and Bontempi, G.}
\newblock minet: {A} {R}/{B}ioconductor {P}ackage for {I}nferring {L}arge
  {T}ranscriptional {N}etworks {U}sing {M}utual {I}nformation.
\newblock {\em BMC bioinformatics 9}, 1 (2008), 461.

\bibitem{Verma92}
{\sc Verma, T., and Pearl, J.}
\newblock An algorithm for deciding if a set of observed independencies has a
  causal explanation.
\newblock In {\em Proceedings of the Eighth international conference on
  uncertainty in artificial intelligence\/} (1992), Morgan Kaufmann Publishers
  Inc., pp.~323--330.

\end{thebibliography}
\end{document}